\documentclass[11pt]{article}

\usepackage[T1]{fontenc}
\usepackage[utf8]{inputenc}
\usepackage[margin=1in]{geometry}
\usepackage{amsmath,amssymb}
\usepackage{booktabs}
\usepackage{graphicx}
\usepackage{microtype}
\usepackage{xcolor}
\usepackage{tikz}
\usetikzlibrary{arrows.meta,positioning,fit,backgrounds}
\usepackage[colorlinks=true,linkcolor=blue,citecolor=blue,urlcolor=blue]{hyperref}

\newcommand{\nats}{\,nats}

\title{Almost Free State Prediction Separation}
\author{John Langford$^{1*}$, Nathan Godey$^{2*}$, Giovanni Monea$^{2}$, Yoav Artzi$^{2}$, Harry Dong$^{1}$, Ying Fan$^{1}$, \\
 Gustavo de Rosa$^{1}$, Zheng Zhan$^{1}$ \\[4pt]
{\small $^{1}$Microsoft \qquad $^{2}$Cornell University \qquad $^{*}$Shared first author}}
\date{}

\begin{document}
\maketitle

\begin{abstract}
State--prediction separation (SPS)~\cite{monea2026sps} relieves a language model's hidden state of two
competing burdens---\emph{summarizing} the context and \emph{predicting} the next token---by splitting the
forward pass into a state stream and a prediction stream. The separation works, but it is expensive: the
prediction stream is a second pass over the whole backbone, costing $\sim$1.9$\times$ the pretraining FLOPs, and even more in terms of wall-clock time when using a flexible attention mask. This paper makes
state--prediction separation almost free. We take the separation to its limit with a \emph{free pause token}:
a prediction stream that writes no keys or values at all and so rides the sequence's existing positions. It
improves next-token prediction of a standard Transformer by 2-3 centinats in practice on a 1B parameter model, and because it adds no
position it costs nothing at inference---no added context length, no KV cache, no decode steps, and
essentially no latency, with the growth in inference flops typically irrelevant as it is not the active
bottleneck on throughput. The cost is therefore entirely in training where we use four mechanisms to drive it down: a
two-pass split that keeps FlashAttention kernels viable, the $w{=}0$ prediction window, a shared gated FFN
that evaluates one FFN per position rather than one per stream, and phasing the separation onto the tail of
the run. Together these bring the overhead versus an optimized pretraining pipeline to $1.33\times$
wall-clock while recovering $\sim$94\% of the gain compared to SPS, and to as low as $1.09\times$ along a graceful
quality/compute tradeoff. Furthermore, the FFN optimization reduces the raw flops required at inference time.  The result is an isoflop, isoparameter, and isotoken improvement over standard next
token trained transformers.
\end{abstract}

\begin{figure}[h!]\centering
\includegraphics[width=0.49\linewidth]{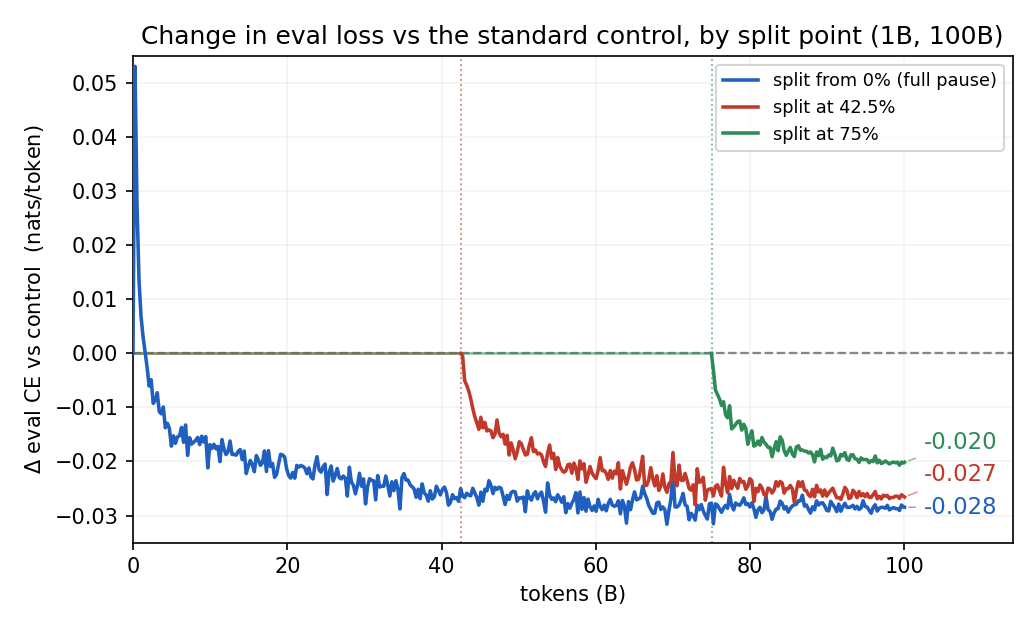}\hfill
\includegraphics[width=0.49\linewidth]{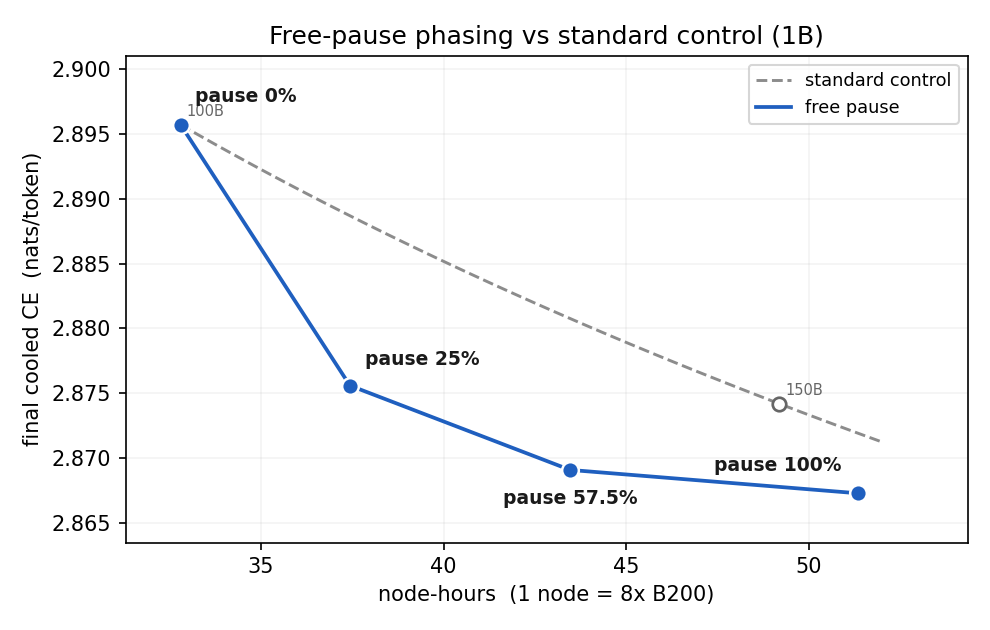}
\caption{\textbf{Left:} change in eval CE against the standard
  control,
  $\Delta(t)=\mathrm{CE}_{\text{variant}}(t)-\mathrm{CE}_{\text{control}}(t)$,
  for three pause phase start points; after each switch the primary
  gain accrues within $\sim$15B tokens with modest further
  gains. \textbf{Right:} compute frontier---final cooled CE
  vs.\ $8\times$B200 node-hours for different pause start
  points; improvements over standard control vary from $-0.005$ to $-0.013$ nats.}
\label{fig:teaser}
\end{figure}

\section{Introduction}

Cross-entropy of next token loss is known to track downstream
capability closely enough to anchor the scaling laws that govern
modern pretraining~\cite{kaplan2020scaling} implying that prediction
and compression are deeply aligned
problems~\cite{shannon1951,deletang2024compression}.  Modern modeling
approaches then spend enormous compute to buy fractional nats per
token, each hard-won and, once folded into a base model, inherited by
everything trained on top.  Given this, even a small but
reliable reduction in next-token loss is valuable provided
it does not cost the parameters, tokens, inference budget, or training
compute that would negate it. Training compute is particularly difficult--an
architectural change that improves loss but doubles the cost of
pretraining is not an improvement at all, because the older architecture training with more tokens may be superior~\cite{hoffmann2022chinchilla}.

A decoder transformer predicts token $x_{i+1}$ from the hidden state $h_i$ at position $i$, produced by
$L$ layers of causal attention and MLPs over the token embeddings. That single state carries two burdens
at once: it must \emph{summarize} $x_{0:i}$---the state that later positions attend to---and simultaneously
\emph{be a good predictor} of $x_{i+1}$. These two goals pull in different directions, yet a model with one
state per position must serve both from the same vector. The \emph{state--prediction separation} (SPS)
hypothesis~\cite{monea2026sps} addresses this tension by giving the two jobs their own streams over a
weight-shared backbone resulting in next-token loss improvements.  This has a substantial training time cost though---a prediction stream is a
second pass over every layer, so SPS asks for roughly $1.9\times$ the pretraining FLOPs of the model it
improves.  Furthermore, wall clock training time may be larger in practice since a flexible mask capable of expressing the SPS solution is not as efficient on a modern GPU.

We take state--prediction separation to its limit with a \emph{free pause token}, which gives the prediction
its own computation without a sequence position. Alongside the ordinary forward pass---the \emph{state} stream $a$, which
summarizes the context and exposes per-layer keys and values---we run a second, weight-shared
\emph{prediction} stream $p$. At every position $p$ starts from one learned embedding, forms a query at each
layer over $a$'s keys and values, and emits the next-token prediction; the training loss falls only on $p$.
Critically, $p$ writes no keys or values of its own, so it adds nothing to the sequence---it rides the
\emph{existing} positions. The state can then specialize in summarizing and the prediction in predicting,
and because the pause occupies no new position the separation is effectively \emph{free at inference}: no added context length, KV
cache, or decode steps, and near-free decode latency (a decode-step microbench on B200; \S\ref{sec:eng}). The primary remaining issue is therefore just \emph{training} cost.

We drive this cost down using four mechanisms, all developed in
\S\ref{sec:method}. First, because the prediction writes no keys
or values, training can instead split into two FlashAttention-friendly~\cite{dao2022flashattention,zadouri2026flashattention4} passes, providing $\sim$4$\times$ throughput over a naive flexible attention mask.  Second, taking the prediction's own attention window to $w{=}0$ avoids a second,
log-sum-exp--merged attention call at only a millinats performance cost. Third, a \emph{shared gated FFN}
evaluates the position-wise FFN once per position rather than once per stream, halving the dominant term in
the second pass's FLOPs and removing its stored activations. Fourth, because the method is iso-parameter, a
standard checkpoint is already a valid backbone, so the separation can be \emph{phased in} for the tail of
the run and the second pass paid on only a fraction of the tokens. Together these take state--prediction
separation from $\sim$1.9$\times$ pretraining FLOPs to $1.33\times$ wall-clock with essentially all of the
gain intact, and to as low as $1.09\times$ if some is traded away.

Section~\ref{sec:related} places this among neighboring lines of work: the free pause delivers the effect of
a pause token~\cite{goyal2023pause} without spending a sequence position on it, and it puts to work the same
spare inference-time compute that speculative
decoding~\cite{cai2024medusa,chen2023accelerating,leviathan2023fast,stern2018blockwise} exploits, for
prediction quality rather than decoding speed.  Everything is evaluated with a
1B scale model on Phi-4 derived pretraining data~\cite{phi4} against a
tightly matched control at global batch 524k; the architecture and
optimization are given in \S\ref{sec:setup}.

Figure~\ref{fig:teaser} previews the payoff at the cheap end of that range. Against the matched control, phasing
the pause onto the run's tail lowers next-token cross-entropy at every point in training
(Fig.~\ref{fig:teaser}, left); plotted against wall-clock node-hours the resulting frontier stays below the
control's own compute-for-loss curve, so the gain survives at equal compute---an iso-compute improvement
(Fig.~\ref{fig:teaser}, right).

\section{Method}
\label{sec:method}

\paragraph{Two weight-shared streams.} Concretely, the state stream $a$ embeds the input tokens and performs causal (optionally sliding-window) self-attention, producing the persistent per-layer keys and values. The prediction stream $p$ carries \emph{no} token embedding: at every position it is initialized from one shared learned vector, \texttt{predict\_embedding}, forms only a query over $a$'s keys and values at each layer, and feeds the LM head. Because every backbone parameter is shared, the model differs from a standard Transformer by exactly one tensor, so a standard checkpoint is already a valid backbone---the property \emph{phasing} exploits below.

\begin{figure}[t]\centering
\begin{tikzpicture}[
  font=\footnotesize, >=Stealth,
  tok/.style   ={rounded corners=1.5pt, draw=black!45, fill=black!7,   minimum width=0.9cm,  minimum height=0.5cm},
  state/.style ={rounded corners=2.5pt, draw=blue!60,  fill=blue!12,   minimum width=1.15cm, minimum height=0.75cm},
  pred/.style  ={rounded corners=2.5pt, draw=orange!85, fill=orange!16, minimum width=1.15cm, minimum height=0.75cm},
  emb/.style   ={rounded corners=2.5pt, draw=orange!85, fill=orange!30, minimum width=1.5cm,  minimum height=0.75cm, align=center},
  flow/.style  ={->, black!55, semithick},
  scol/.style  ={->, blue!65, semithick},
  pcol/.style  ={->, orange!90, semithick},
  read/.style  ={->, orange!90, densely dashed, semithick},
]
\def\dx{3.0}
\foreach \i/\lab/\olab in {0/{i{-}1}/{i},1/{i}/{i{+}1},2/{i{+}1}/{i{+}2}}{
  \node[tok]   (x\i) at (\i*\dx,0)    {$x_{\lab}$};
  \node[state] (a\i) at (\i*\dx,1.45) {$a_{\lab}$};
  \node[pred]  (p\i) at (\i*\dx,3.15) {$p_{\lab}$};
  \draw[flow] (x\i) -- (a\i);
  \draw[pcol] (p\i) -- ++(0,1.0);
  \node[anchor=south] at (\i*\dx,4.15) {$\hat{x}_{\olab}$};
}
\begin{scope}[on background layer]
  \node[rounded corners=4pt, fill=blue!5,   fit=(a0)(a2), inner xsep=10pt, inner ysep=7pt] (Sbox){};
  \node[rounded corners=4pt, fill=orange!6, fit=(p0)(p2), inner xsep=10pt, inner ysep=7pt] (Pbox){};
\end{scope}
\node[blue!60,   anchor=west, font=\footnotesize\itshape] at (Sbox.east) {\,state};
\node[orange!90, anchor=west, font=\footnotesize\itshape] at (Pbox.east) {\,prediction};
% state: causal self-attention writes + reads K/V
\draw[scol] (a0) -- node[below=1pt, font=\scriptsize, black!55]{K,V} (a1);
\draw[scol] (a1) -- (a2);
% shared pause embedding seeds the prediction stream at every position
\node[emb] (ep) at (-2.55,3.15) {$e_{\text{pause}}$};
\draw[pcol] (ep) -- (Pbox.west);
% prediction forms a query over the state's K/V, writing none
\foreach \i in {0,1,2}{ \draw[read] (a\i) -- (p\i); }
\node[anchor=south east, black!70, font=\scriptsize] at (-1.9,4.05) {loss $\uparrow$};
% legend
\node[anchor=west, black!75, font=\scriptsize, align=left] at (-3.3,-0.95)
 {solid $=$ writes/uses K/V \quad\ \textcolor{orange!90}{dashed} $=$ query only, no K/V written \quad\ $p_i$ attends $a_{\le i}$ \quad\ weights shared, $\times L$ layers};
\end{tikzpicture}
\caption{The free pause at one layer, over three positions. The \textcolor{blue!60}{\emph{state}} stream $a$ is
the ordinary causal pass and writes the per-layer keys and values. The \textcolor{orange!90}{\emph{prediction}}
stream $p$ is initialized at every position by the same shared embedding $e_{\text{pause}}$, forms only a
\emph{query} over the state's keys and values (dashed; causally $p_i$ reads $a_{\le i}$), and writes none of its
own; its output $\hat{x}$ is scored by the next-token loss. The streams share all weights and this repeats over
$L$ layers. Because $p$ adds no key/value and no sequence position, the pause is free at inference; the only
added parameter is $e_{\text{pause}}$.}
\label{fig:method}
\end{figure}
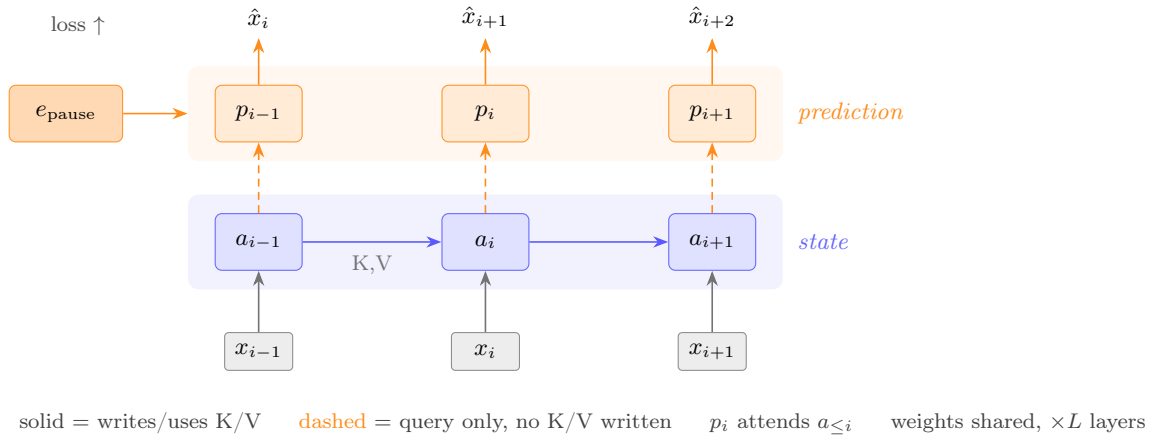

\paragraph{A FlashAttention-friendly two-pass split.}
During training, the two passes are ordered: the state stream runs first to produce the cached keys and values, then the prediction stream runs as a plain cross-attention over them (predict query $\times$ state key/value). Both are shapes that FlashAttention kernels~\cite{dao2022flashattention} express directly, whereas the interleaved form's mask---neither causal, sliding-window, nor block-diagonal over packed sequences---is not. We run the Blackwell-targeted FlashAttention-4~\cite{zadouri2026flashattention4} on B200. We measure 73k tokens/s per GPU instead of 16k tokens/s with a single pass flexible attention mask, about a 4$\times$ improvement.

\paragraph{A shared gated FFN.} Since the majority of an LLM's parameters reside in the FFNs they require heavy computation.  Given this, it may be  desirable to join these computations for the state and prediction streams. Let $\bar a$
and $\bar p$ be the post-norm state and prediction residuals entering the FFN sub-layer. A per-token scalar
gate pools them, a single FFN is applied to the pooled input, and two further scalar gates route its output
back into each stream:
\[
g=\sigma\!\left(W_{\text{in}}[\bar a;\bar p]\right),\qquad
f=\mathrm{FFN}\!\left(g\,\bar a+(1{-}g)\,\bar p\right),\qquad
a\mathrel{+}=\sigma(W_a \bar a)\,f,\quad
p\mathrel{+}=\sigma(W_p \bar p)\,f .
\]
One FFN evaluation per position thus replaces two. The three added gates are
single-output linears (negligible parameters), zero-initialized so that every gate starts at $0.5$. No
backbone parameter changes, so a standard checkpoint can still be \emph{phased in}. The switch is not fully
function-preserving, however: at initialization the sub-layer adds
$\tfrac{1}{2}\mathrm{FFN}(\tfrac{1}{2}\bar a+\tfrac{1}{2}\bar p)$ to both streams, rather than
$\mathrm{FFN}(\bar a)$ to the state and $\mathrm{FFN}(\bar p)$ to the prediction, so unlike the two-pass
form it requires some training to settle after the switch. The FLOP, memory, and quality consequences are
measured in \S\ref{sec:sharedffn}.

\paragraph{Phasing.} Training runs as for a standard next token prediction transformer for a fraction $f$ of the schedule and switches on the split for
the remainder, paying the second pass on only a $1-f$ token fraction (compute $f + (1-f)\cdot 1.57$).  The free pause training phase starts from standard training with all training state intact (step, schedule, backbone optimizer, and dataloader all continue;
only \texttt{predict\_embedding} initializes fresh).  In practice, this phase switchover works well with immediate evaluation loss improvements.

\section{Related work}
\label{sec:related}

Related work spans state/prediction separation, pause (aka thinking) tokens, and speculative decoding.

\paragraph{State--prediction separation.} \cite{monea2026sps} introduced the hypothesis this paper builds on
and the architecture that tests it: the forward pass is split into a state stream and a prediction stream so
that summarizing and predicting need not share one representation. In their formulation the prediction stream
still writes keys and values that are retained within a sliding window of $w$ tokens. Our free pause is the
$w{=}0$ limit taken strictly: the prediction writes nothing at all, forming
only a query over the state's keys and values (differently from SPS $w{=}0$, which also writes a temporary key and value for the prediction stream). This stricter separation is what makes the method free at
inference since the prediction adds no cache.  It is also what makes the training cheap: with no prediction
keys or values, pure FlashAttention works in training and the prediction reduces to a plain cross-attention
pass (\S\ref{sec:method}). The shared gated FFN and the phasing schedule then drop the computational cost of
using this in pretraining to a small and clearly viable tradeoff.

\paragraph{Pause and thinking tokens.} \cite{goyal2023pause} adds computation before a prediction by
inserting learned, non-vocabulary \emph{positions} into the sequence, giving the model extra forward passes
to ``think'' before it commits. The free pause supplies the same extra per-position computation, but on a
parallel stream rather than a new position: a pause \emph{token} occupies its own position and so enlarges
the context, the KV cache, and the number of decode steps, whereas the free pause leaves all three
unchanged.

\paragraph{Speculative and parallel decoding.} Speculative decoding and related parallel-decoding
methods~\cite{stern2018blockwise,leviathan2023fast,chen2023accelerating,cai2024medusa} also advance more
than one token-query within a single decode step. There the extra queries are \emph{speculative} future
continuations that a verifier accepts or rejects; the free pause's second query is instead a prediction at
the \emph{current} position that is always kept, and it improves prediction quality rather than speeding up
generation.  A free pause token leverages exactly the same spare compute which speculative decoding benefits from for the purpose of improving prediction quality rather than prediction speed.  We leave investigating the combination of both techniques here to later work.

\paragraph{Parallel efforts.} Decode-Branch Transformers~\cite{liu2026decodebranchtransformersdecouplingprimary} aims at a similar idea.  There are of course differences in the details, and qualitatively in the phasing optimization here and with the broader study of Mixture of Experts models there.  Overall, these results reinforce the value and scope of free pause tokens.

\section{Model and optimization}
\label{sec:setup}

\paragraph{Architecture.} A 1B decoder transformer: 24 layers, hidden size $1536$,
grouped-query attention~\cite{ainslie2023gqa} with $16$ query and $8$ key value heads, sliding-window attention~\cite{beltagy2020longformer,jiang2023mistral} (window $2048$)
on most layers with a periodic full-attention layer every 6 layers~\cite{gemma2}, QK-normalization~\cite{henry2020qknorm}, partial rotary embeddings~\cite{su2021roformer,black2022neox} on the
full-attention layers, and tied input/output embeddings~\cite{press2017tied}; sequence length $8192$.

\paragraph{Optimization.} A Muon-family optimizer~\cite{jordan2024muon} at peak learning rate $2\mathrm{e}{-}2$ on a
warmup--stable--cooldown schedule~\cite{hu2024minicpm,hagele2024scaling} (a short warmup, a constant plateau, then a final-25\% linear cooldown to
$2\mathrm{e}{-}3$). Global batch $524{,}288$ tokens (micro-batch 4 with gradient-accumulation 2 across
$8\times$B200), bf16 activations with mxfp8~\cite{rouhani2023microscaling} matmuls.

\paragraph{Data and baseline.} Pretraining on Phi-4 derived data~\cite{phi4}. The strong baseline is the identical
model with the prediction stream removed, matched on optimizer, data order, schedule, and global batch, so
the two differ only in the second pass---and hence in wall-clock throughput.

\section{Measurement}
\label{sec:methodology}

The prediction pass reruns the transformer layer stack but grafts the state's key value pairs, skipping the key value projections implying a cost of 
$\sim$1.9$\times$ FLOPs, not $2\times$. Its wall-clock overhead is lower still ($\sim$1.57$\times$),
because the compute-dense pass runs at $\sim$20\% higher utilization than the micro-batch-4 baseline. Compute is reported as
\emph{wall-clock node-hours} on identical $8\times$B200 hardware---the cost actually paid.  Hence, \emph{iso-compute}
below means iso-node-hours which account for the higher utilization.  At true iso-FLOP ($\sim$1.9$\times$) the
control has more tokens and the margins tighten---the full pause turns slightly negative and the phased deltas
roughly halve, though the ordering is unchanged. 

\section{Results}
\label{sec:results}

With a strong baseline (global batch 524k), the free
pause reaches 2.8673 versus the control's 2.8957---a $-0.0284$\nats\ iso-token gain.
We measure each mechanism of \S\ref{sec:method} against the second pass's cost, and then ask whether
the gain survives once the control is handed the compute the separation would have consumed.

\subsection{Cutting the second pass: a FlashAttention-friendly split, $w{=}0$, the shared FFN, and phasing}
\label{sec:levers}

\begin{table}[h]\centering\small
\begin{tabular}{lll}
\toprule
change & effect on cost & quality \\
\midrule
FA-friendly split & interleaved $\sim$4$\times$ $\to$ the $1.57\times$ baseline & --- \\
$w{=}0$ & avoids the window's $1.22\times$ pass & costs $\le0.009$ ($-0.0047$@100B) \\
shared gated FFN & $1.57\times \to 1.35\times$ & costs $\sim$0.005--0.010 \\
phasing (42.5\%) & $1.57\times \to 1.33\times$ & recovers $\sim$all (2.8691 vs 2.8673) \\
\bottomrule
\end{tabular}
\caption{Each change targets the second-pass cost. Overheads are wall-clock, relative to the strong
baseline ($1.00\times$).}
\label{tab:levers}
\end{table}

\textbf{A FlashAttention-friendly split.} Reorganizing the computation into two distinct stream passes---rather than one interleaved attention---is what lets stock fused kernels run the model at all, and it makes the whole run much faster.

\textbf{Eliminate prediction-only attention.} The original SPS paper~\cite{monea2026sps} has an additional small sliding window amongst the prediction key/values.  Here we find that the more extreme $w{=}0$ choice is a reasonable since the prediction
self-window helps by only $-0.0047$ with 100B tokens, and $w{>}0$
needs a second, log-sum-exp--merged FA call increasing compute by $1.22\times$ the $w=0$ pass (Sec.~\ref{sec:eng}).  Thus using 
archive-only key values is the fast default. 

\textbf{Shared gated FFN.} Evaluating the position-wise FFN once per position rather than once per stream
removes about half of the second pass's dominant term, taking a full pause from $1.57\times$ to $1.35\times$
wall-clock and freeing enough memory for a faster micro-batch. It gives up $\sim$0.005--0.010 nats against the
two-pass form. \S\ref{sec:sharedffn} reports the memory, throughput, and quality measurements.

\begin{figure}[h]\centering
\includegraphics[width=0.74\linewidth]{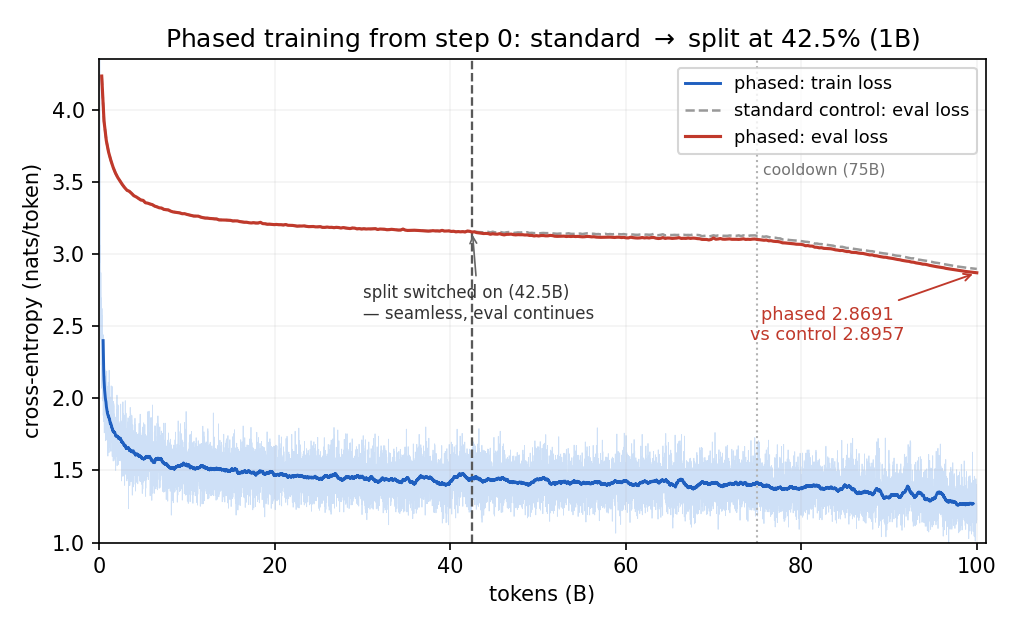}
\caption{Phased training (42.5\% split) stitched from step 0: standard pretraining to 42.5B, then the
split to 100B. Eval loss (red) overlays the standard control (dashed) up to the switch---seamless, no
spike---then diverges below it; cooldown at 75B brings the phased run to $2.8691$ versus the control's
$2.8957$. Train loss (blue) is the pretraining loss.}
\label{fig:phasedloss}
\end{figure}

\textbf{Phasing.}
Switching to the pause at 42.5\% recovers essentially all the gain (2.8691, within noise of full pause's
2.8673) at $1.33\times$ instead of $1.57\times$ (Table~\ref{tab:phased}). The 75\% split ($1.14\times$)
tests the hard case where the pause gets only the cooldown to adapt. Stitched from step 0
(Fig.~\ref{fig:phasedloss}), the phased run tracks the control's loss up to the switch---the switch is
seamless, no visible spike---then diverges below it and cools to $2.8691$. The CE-vs-node-hours frontier
(Fig.~\ref{fig:teaser}, right) is convex: a short pause finish banks most of the gain per extra node-hour.
Plotting the change in eval loss against the control for different free pause phase points (Fig.~\ref{fig:teaser}, left)
shows that a cold start with the full pause is actually worse for training of $\sim$1.5B tokens but then provides a clear win either continuing or with phases starting later.   Most of the benefit accrues within $\sim$15B tokens of free pause training although small gains continue to be observed the longer free pause training continues.  Given this structure, late phase free pause token training achieves most of the gains of early phases for a small fraction of the overall compute.

\begin{table}[h]\centering\small
\begin{tabular}{lrrl}
\toprule
schedule & 100B CE & compute & vs full pause \\
\midrule
control (no pause)     & 2.8957 & $1.00\times$ & --- \\
full pause ($w=0$)     & 2.8673 & $1.57\times$ & --- \\
42.5\% std $\to$ pause & 2.8691 & $1.33\times$ & $+0.0018$ (recovers 94\%) \\
75\% std $\to$ pause   & 2.8756 & $1.14\times$ & $+0.0083$ (recovers 71\%) \\
\bottomrule
\end{tabular}
\caption{Phasing pays the pass only on the tail; full-resumed from a standard iso-parameter checkpoint.}
\label{tab:phased}
\end{table}

\subsection{The shared gated FFN in practice}
\label{sec:sharedffn}

The shared gated FFN (\S\ref{sec:method}) evaluates the position-wise FFN once per position instead of once
per stream. What that buys is measured here.

\paragraph{Memory and throughput.} At sequence length 8192 the shared form removes the second pass's stored
FFN activations---about $14$ GB/GPU less peak memory ($91$ vs $105$ GB at micro-batch 4), enough that it fits
a larger, faster micro-batch on which the two-pass form OOMs. On the same hardware the shared-FFN pause then
runs at $0.74\times$ the control's tokens/s/GPU ($\sim$95k vs $\sim$128k), versus $0.64\times$ for the
two-pass form---closing roughly a third of the free pause's throughput penalty. A full pause therefore costs
$1.35\times$ control wall-clock rather than $1.57\times$, and phased at 75\% it costs $1.09\times$---the
cheapest point in the paper (Table~\ref{tab:sharedffn}).

\begin{figure}[h]\centering
\includegraphics[width=0.49\linewidth]{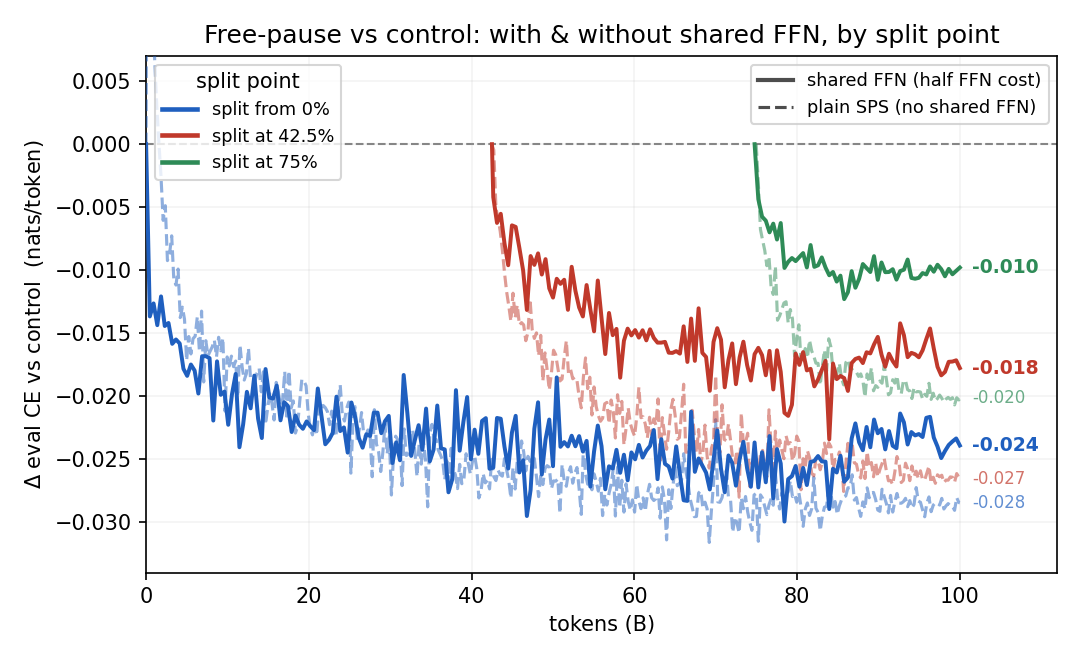}\hfill
\includegraphics[width=0.49\linewidth]{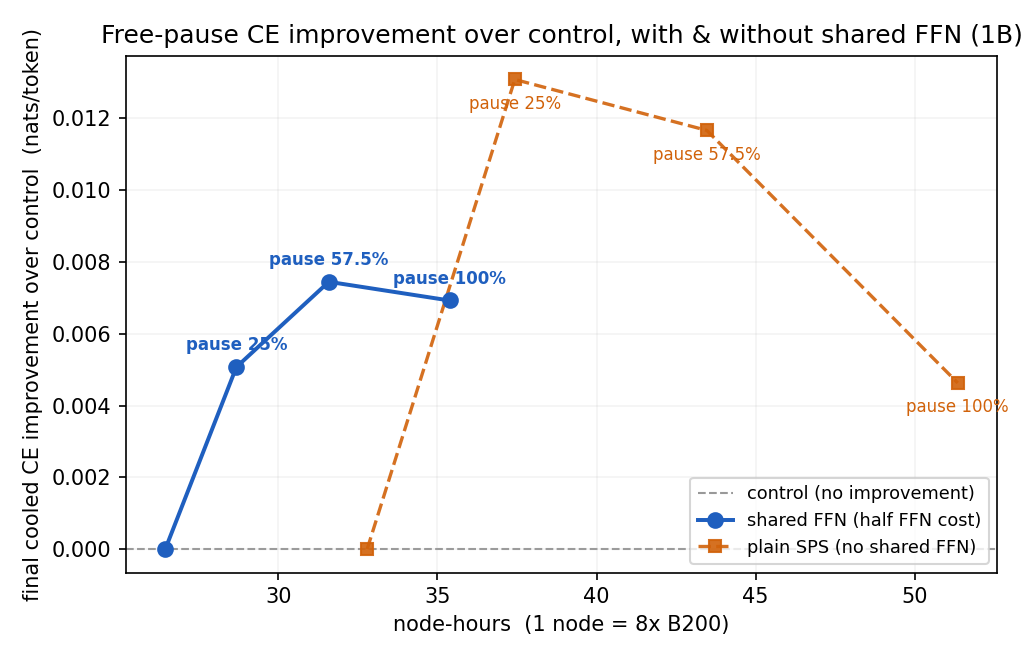}
\caption{Shared gated FFN vs the two-pass free pause. \emph{Left:} change in eval CE against control by split
point; solid $=$ shared FFN, dashed $=$ two-pass. The shared form keeps most of the advantage at half the FFN
cost. \emph{Right:} final cooled CE improvement over each variant's \emph{own} control vs node-hours---this
normalizes out the different baselines. The shared-FFN frontier (blue) reaches its gain at far fewer node-hours
than the two-pass form (orange).}
\label{fig:sharedffn}
\end{figure}

\paragraph{Quality.} Trained fresh or phased in, the shared-FFN pause still beats the control at every split
point (Table~\ref{tab:sharedffn}, Fig.~\ref{fig:sharedffn} left): $-0.0239$ from a cold start, $-0.0178$ at a
42.5\% split, $-0.0098$ at 75\%. It gives up some of the two-pass form's advantage ($\sim$0.005--0.010
nats/token), so there is a real cost associated with the shared FFN. Part of that apparent gap is a baseline artifact: to fit
the faster micro-batch the shared runs use a slightly smaller global batch whose own control is $\sim$0.011
nats/token worse, so the iso-batch cost of sharing is smaller than the raw deltas suggest. Normalizing each
variant to \emph{its own} control (Fig.~\ref{fig:sharedffn} right) puts the shared-FFN frontier well to the left
of the two-pass one: it reaches its improvement at far fewer node-hours.

\begin{table}[h]\centering\small
\begin{tabular}{lrrr}
\toprule
schedule (shared FFN) & 100B CE & $\Delta$ vs control & compute \\
\midrule
control (no pause)     & 2.9064 & ---       & $1.00\times$ \\
full pause ($w=0$)     & 2.8825 & $-0.0239$ & $1.35\times$ \\
42.5\% std $\to$ pause & 2.8886 & $-0.0178$ & $1.20\times$ \\
75\% std $\to$ pause   & 2.8966 & $-0.0098$ & $1.09\times$ \\
\bottomrule
\end{tabular}
\caption{Shared gated FFN: one FFN evaluation per position instead of two. Throughput $0.74\times$ control (vs
$0.64\times$ for the two-pass form); still beats control at every split point. Measured at the faster micro-batch
operating point, so the control is the matched shared-batch control ($2.8957\to2.9064$; see text).}
\label{tab:sharedffn}
\end{table}

\subsection{Iso-FLOP analysis}

At iso-compute the baseline spends the saved node-hours on extra tokens; the comparison is the pause at
100B against the control given the same node-hours (Table~\ref{tab:isoflop}). The control is measured to
both 100B and 150B, so its cooled loss falls at a measured $\sim$0.037\nats/doubling. At equal node-hours
\emph{every} free-pause variant is ahead with shortest free phase giving a $-0.013$ advantage and full pause training providing only a $-0.005$ advantage.  The margins tighten with a stricter iso-FLOP ($\sim$1.9$\times$) analysis where the phased runs stay
positive ($-0.005$ to $-0.009$) while the full pause turns slightly negative ($+0.006$).  Overall, free pause token training towards the end of pretraining is a clear and desirable compute win of $\sim$1 centinat depending on how measurements are done.  Note the ordering: the \emph{cheapest} schedules are the ones that win most at equal compute, which is precisely the point of making the separation almost free rather than merely effective.

\begin{table}[h]\centering\small
\begin{tabular}{lrrll}
\toprule
variant & CE (100B) & compute & baseline @ iso-compute & $\Delta$ \\
\midrule
full pause ($w=0$)     & 2.8673 & $1.57\times$ & ctrl @ 157B $=$ 2.872 & $-0.005$ \\
42.5\% std $\to$ pause & 2.8691 & $1.33\times$ & ctrl @ 133B $=$ 2.881 & $-0.012$ \\
75\% std $\to$ pause   & 2.8756 & $1.14\times$ & ctrl @ 114B $=$ 2.889 & $-0.013$ \\
\bottomrule
\end{tabular}
\caption{Iso-compute (node-hours) comparison against the strong baseline, from the control's
\emph{measured} 100B and 150B cooled endpoints (slope $\sim$0.037\nats/doubling): 114--133B interpolated,
157B a short extrapolation. Every free-pause variant is ahead at equal node-hours.}
\label{tab:isoflop}
\end{table}

\subsection{Inference cost}
\label{sec:inference}

At input prefill, only $x_i$ need to be forwarded (no prediction is required and no keys and values are made by $p_i$) so the cost is the same as a vanilla Transformer. At decode, $(x_i, p_i)$ forward together as a two-token step: the state
attention appends $x_i$'s key value pairs to the cache and the prediction reads it, the two co-advancing layer by layer.
It is very typical for autoregressive latency to be set by the decode's sequential depth (one step per generated token, $L$
layers each), which the free pause leaves unchanged. It adds only the prediction stream's \emph{parallel}
compute within each step---extra FLOPs, not extra depth. With a small batch decode that is typically
hidden, so batching the two streams' projections and sharing the KV read across the two queries put the fused
step within $\sim$1\% of a standard decode (\S\ref{sec:eng}).

\subsection{Downstream evaluation}

Does the per-token gain survive as downstream capability, or is it loss-only? We evaluate the cooled
checkpoints on two aggregate measures whose sampling error is small enough to resolve differences at 1B:
DCLM CORE~\cite{li2024dclm} (a centered mean over 21 multiple-choice tasks) and held-out bits-per-byte
(Table~\ref{tab:downstream}). Individual lm-eval~\cite{gao2023lmeval} task scores are within their 95\% confidence intervals so relevant signal is in these pooled and dense measures. Iso-token (both 100B), the full pause
raises DCLM CORE from $0.327$ to $0.347$ and lowers climbmix~\cite{diao2025climb} BPB by $0.008$, providing modest benefit consistent \emph{with} the $-0.028$\nats\ CE gain. Iso-compute, the pause
at 100B matches the control given 50\% more tokens (150B) to within measurement error on both metrics, and
phasing reaches the same level at $1.14$--$1.33\times$ compute. The gain is therefore not a loss-only
artifact: it appears in downstream compression and in aggregate task accuracy.

\begin{table}[h]\centering\small
\begin{tabular}{lrr}
\toprule
variant & DCLM CORE $\uparrow$ & BPB climbmix $\downarrow$ \\
\midrule
control, 100B              & 0.327 & 0.777 \\
control, 150B              & 0.348 & 0.770 \\
full pause ($w{=}0$), 100B & 0.347 & 0.769 \\
42.5\% $\to$ pause, 100B   & 0.348 & 0.769 \\
75\% $\to$ pause, 100B     & 0.348 & 0.771 \\
\bottomrule
\end{tabular}
\caption{Downstream metrics at 1B: DCLM CORE (centered mean over 21 tasks; SE $\approx0.01$) and held-out
bits-per-byte on climbmix (SE $\approx0.004$), all at sequence length 8192.
Iso-token, the full pause beats the 100B control on every aggregate; iso-compute it matches the 150B
control, and the phased schedules reach the same level for far less compute. Per-task lm-eval scores are
individually within their 95\% intervals and are not shown.}
\label{tab:downstream}
\end{table}

\subsection{What the pause embedding learns}

\begin{figure}[h]\centering
\includegraphics[width=0.72\linewidth]{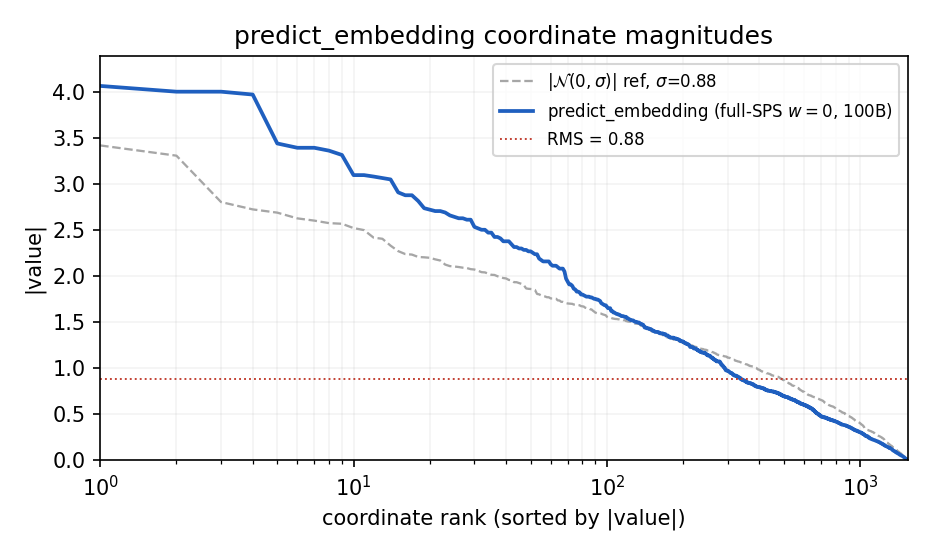}
\caption{Sorted coordinate magnitudes of the learned pause embedding (the full pause, $w{=}0$, 100B), against a
Gaussian reference at the same RMS ($0.88$). A handful of channels sit well above Gaussian ($\pm3$--$4$,
$\approx4\sigma$; top-10 dims $=11\%$ of the energy) over a near-Gaussian unit-RMS bulk.}
\label{fig:pe}
\end{figure}

The single added embedding grows $\sim$40$\times$ from init to
per-coordinate RMS $\approx1$ ($\|p\|\approx\sqrt{d}$, $\sim$0.4$\times$ a token-embedding norm), but is
RMS-normalized before use, so only its direction matters. That direction leans modestly toward the
frequency prior.  The nearest tokens by cosine are the commonest continuations (comma,
period, newline, `the', `and'; cosine $0.25$--$0.46$). A
dozen-odd coordinates reach $\pm3$--$4$ ($\approx4\sigma$; Fig.~\ref{fig:pe}), carrying $\sim$11\% of the
energy over a diffuse bulk. Against the token-embedding table it is a distinct, atypical point: its norm
sits below the used-token shell (22nd percentile of a bimodal norm distribution), it is near-orthogonal to
the embeddings' strong common mode (cosine $0.05$, below 96\% of tokens), and its spikes fall on
idiosyncratic channels rather than the model's high-variance (massive-activation~\cite{sun2024massive}) channels (top-16 overlap
$2/16$). 

\section{Conclusion}

Our experiments here are limited to a single scale (1B) with a single primary seed.  However, in our experience the pretraining process is robust enough that the results reported here are beyond the noise level.  Within that scope, state--prediction separation
provides a modest iso-token/parameter/compute/training FLOP advantage over standard pretraining
methodology. The $\sim$1.9$\times$
pretraining FLOPs that made the separation an expensive curiosity fall to $1.33\times$ wall-clock with
essentially all of the gain intact, and to as low as $1.09\times$ if some is traded away, while inference
stays free. One perhaps-significant variation that may matter in practice is combining this approach with
multitoken prediction~\cite{gloeckle2024multitoken}.

\bibliographystyle{plain}
\bibliography{references}

\appendix
\section{Engineering}
\label{sec:eng}
\textbf{vLLM serving \& decode latency.} A two-stream vLLM~\cite{kwon2023vllm} model serves the free pause: the
  state stream writes paged KV and the prediction stream reads the same layer's KV read-only via cross-layer
  KV-sharing (no extra cache), with all weights shared and both streams run as one $2T$-row batch so each
  layer's weights load once. Sharing the KV read across the two queries---folding them into a single paged
  \texttt{flash\_attn\_with\_kvcache} call (archive-only write, both attend the shared KV)---puts the fused
  free-pause decode step within $\sim$1\% of a standard decode in a decode-step microbench on B200. 

\section{Cross-entropy summary}
\begin{table}[h]\centering\small
\begin{tabular}{lr}
\toprule
role & CE \\
\midrule
control 100B (baseline) & 2.8957 \\
control 150B (extended) & 2.8742 \\
pause $w{=}0$ 100B      & 2.8673 \\
pause $w{=}64$ 100B     & 2.8626 \\
phased 42.5\%           & 2.8691 \\
phased 75\%             & 2.8756 \\
\bottomrule
\end{tabular}
\end{table}

\end{document}